%% file: main.tex
\documentclass[10pt,twocolumn,letterpaper]{article}

\input{preamble}

 \usepackage[pagenumbers]{cvpr} 

\definecolor{cvprblue}{rgb}{0.21,0.49,0.74}
\usepackage[pagebackref,breaklinks,colorlinks,citecolor=cvprblue]{hyperref}

\def\paperID{197} 
\def\confName{3DV\xspace}
\def\confYear{2024\xspace}

\title{Towards Learning Monocular 3D Object Localization From 2D Labels Using the Physical Laws of Motion}

\author{
Daniel Kienzle \hspace{1cm} Katja Ludwig \hspace{1cm} Julian Lorenz \hspace{1cm} Rainer Lienhart \\
University of Augsburg\\
86159 Augsburg, Germany\\
{\tt\small \{firstname.lastname\}@uni-a.de}
}

\begin{document}
\maketitle
\input{sec/0_abstract}

\input{sec/1_intro}


\begin{figure*}
  \centering
  \begin{subfigure}[t]{0.63\linewidth}
    \centering
    \includegraphics[width=\linewidth]{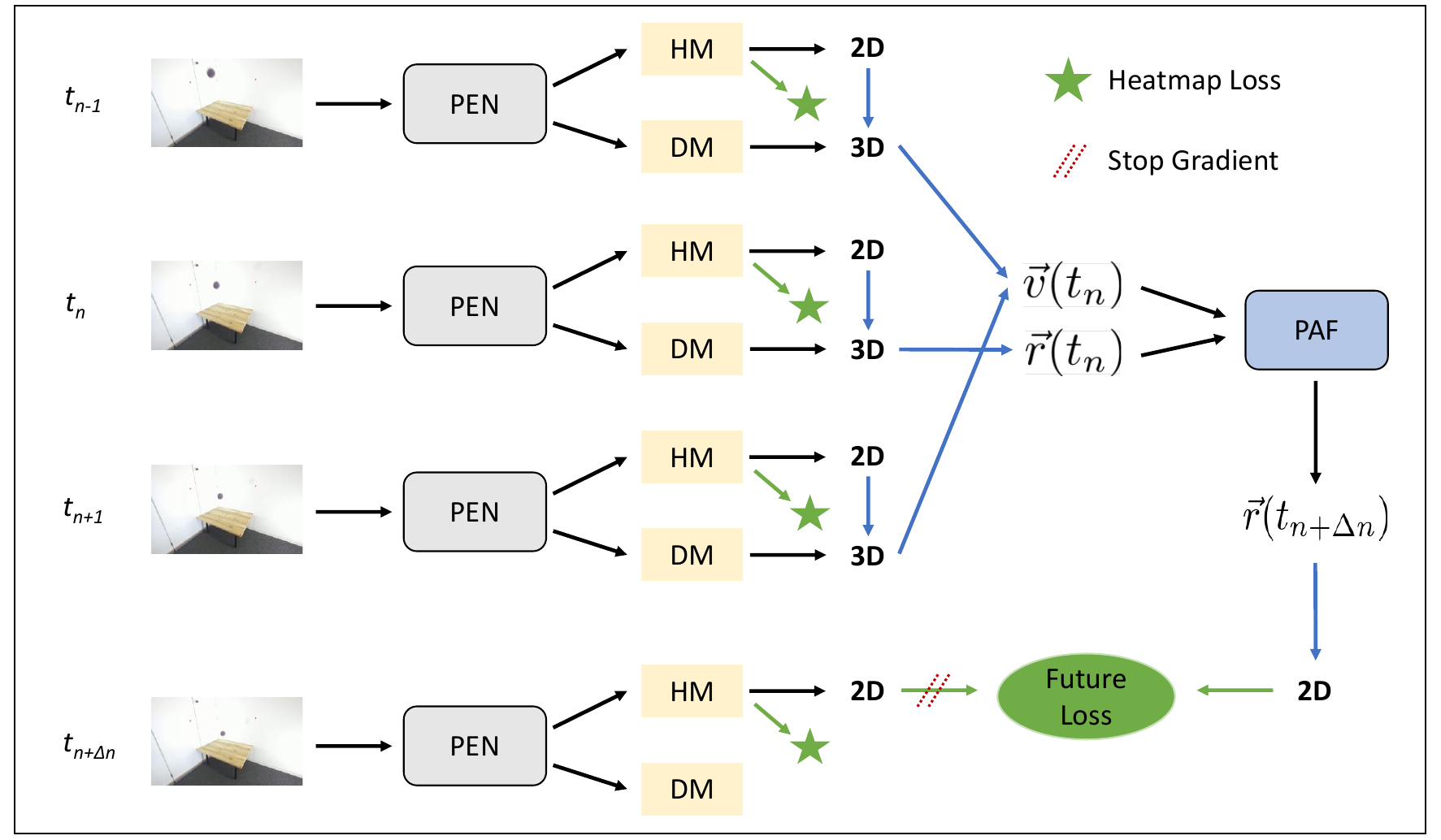}
    \subcaption{Training with videos using explicit physical knowledge.}
  \end{subfigure}
  \hfill
  \begin{subfigure}[t]{0.212\linewidth}
    \centering
    \includegraphics[width=\linewidth]{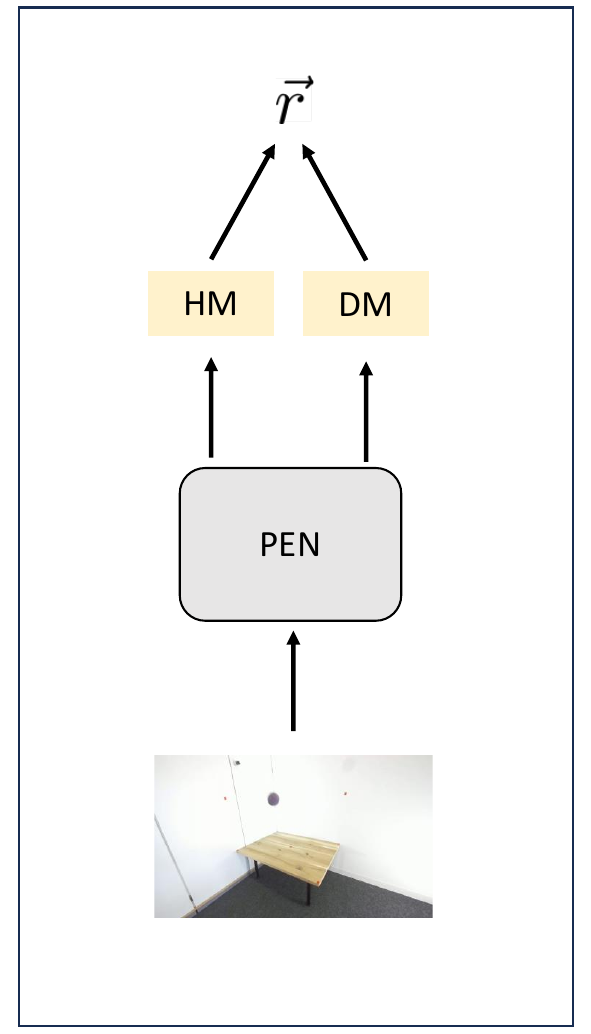}
    \captionsetup{width=\textwidth}
    \subcaption{Inference with single images. No explicit physical knowledge necessary.}
  \end{subfigure}
  \vspace{-0.1cm}
  \caption{Training and inference pipeline. The \textit{Position Estimation Network (PEN)} estimates the heatmap (HM) and depthmap (DM) for each input image, from which the 2D image coordinates and camera depth are extracted. Using the intrinsic camera matrix, we compute the 3D camera coordinates from the image coordinates and camera depth. With the extrinsic camera matrix, the world coordinates \( \vec{r} \) are calculated. During training, we apply the \textit{PEN} to multiple coherent frames for the estimation of the velocity \( \vec{v} \). The \textit{Physics Aware Forecast Module (PAF)} utilizes these initial conditions to calculate the world coordinates at a later time \( t_{n + \Delta n} \) by solving the differential equations of motion. These coordinates are then projected back to image coordinates and the \textit{future loss} is calculated. The \textit{heatmap loss} is calculated by comparing the predicted and ground truth heatmaps at all time steps. During inference, the PEN is applied to single images. Thus, it is robust to incalculable situations like sudden wind or human interventions. Blue arrows indicate the use of camera matrices, and dotted red lines indicate detached gradients.
       }
  \label{img:overview}
  \vspace{-0.3cm}
\end{figure*}

\input{sec/2_relatedwork}

\input{sec/3_method}

  \begin{table*}[th]
    \centering
    \small
    \caption{\( \textit{DtG} \) scores for the \textit{synthetic dataset}.}
    \vspace{-0.1cm}
    \label{tab:DtG_synthetic}
    \begin{subtable}{0.45\linewidth}
    \captionsetup{width=0.9\textwidth}
    \caption{\( \textit{DtG} \) scores per camera location evaluated on the 1\textsuperscript{st} environment.}
    \label{tab:DtG_synthetic_cam}
    \centering
    \resizebox{0.9\linewidth}{!}{
    \begin{tabular}{ccccc}
      \toprule
       & \multicolumn{4}{c}{ \( \textit{DtG} \pm \Delta \textit{DtG} \) (cm) \(\downarrow\)} \\
      \cmidrule(r){2-5} 
      training set & camera 1 & camera 7 & camera 8 & camera 9 \\
      \midrule
      \textit{SD-S} & \( 22 \pm 19 \) & - & - & - \\
      \textit{SD-M}  & \( 19 \pm 10 \) & \( 27 \pm 23 \) & \( 23 \pm 9 \) & \( 21 \pm 10 \) \\
      \textit{SD-L} & \( 11 \pm 6 \) & \( 28 \pm 25 \) & \( 15 \pm 8 \) & \( 16 \pm 7 \) \\
      \bottomrule
    \end{tabular}
    }
    \end{subtable}
    \hfill
    \begin{subtable}{0.35\linewidth}
    \small
    \captionsetup{width=0.9\textwidth}
    \caption{\( \textit{DtG} \) scores per environment.}
    \label{tab:DtG_synthetic_env}
    \centering
    \resizebox{0.9\linewidth}{!}{
    \begin{tabular}{cccc}
      \toprule
       & \multicolumn{3}{c}{ \( \textit{DtG} \pm \Delta \textit{DtG} \) (cm) \(\downarrow\)} \\
      \cmidrule(r){2-4} 
      training set & env 1 & env 2 & env 3  \\
      \midrule
      \textit{SD-S} & - & - & - \\
      \textit{SD-M}  & \( 22 \pm 16 \) & - & - \\
      \textit{SD-L} & \( 18 \pm 16 \) & \( 17 \pm 16 \) & \( 18 \pm 13 \) \\
      \bottomrule
    \end{tabular}
    }
    \end{subtable}
    \normalsize
    \vspace{-0.3cm}
  \end{table*}

\input{sec/4_dataset}

\input{sec/5_experiments}

\FloatBarrier
\input{sec/6_conclusion}

\newpage

{
    \small
    \bibliographystyle{ieeenat_fullname}
    \bibliography{main}
    
}

\input{sec/X_suppl}

\end{document}

%% file: preamble.tex
\usepackage[dvipsnames]{xcolor}

\usepackage[pdftex]{graphicx}
\usepackage{amsmath}
\usepackage{physics}
\usepackage{booktabs}
\usepackage{algorithm}
\usepackage{algorithmic}
\usepackage{stfloats}
\usepackage{siunitx}
\usepackage{subcaption}
\usepackage{placeins}
\usepackage[english]{babel}
\usepackage{wrapfig}
\usepackage{comment}
\usepackage{mathtools}
\usepackage{paralist}
\usepackage{enumitem}

\DeclareSIUnit{\fps}{ \text{FPS} }

%% file: sec/0_abstract.tex
\begin{abstract}
    We present a novel method for precise 3D object localization in single images from a single calibrated camera using only 2D labels. No expensive 3D labels are needed. Thus, instead of using 3D labels, our model is trained with easy-to-annotate 2D labels along with the physical knowledge of the object's motion. Given this information, the model can infer the latent third dimension, even though it has never seen this information during training. Our method is evaluated on both synthetic and real-world datasets, and we are able to achieve a mean distance error of just \( \SI{6}{cm} \) in our experiments on real data. The results indicate the method's potential as a step towards learning 3D object location estimation, where collecting 3D data for training is not feasible.
\end{abstract}

%% file: sec/1_intro.tex
\vspace{-0.3cm}

\section{Introduction}
\label{sec:intro}

The 3D location of objects is crucial in many application domains, such as in robotics or in the performance analysis of athletes in sports tournaments. Especially sports broadcasting companies are greatly interested in obtaining 3D information about an object's location in a fixed environment \cite{AccurateBallTrajectoryTrackingSportsBroadcast}, for example the position of the ball in soccer, basketball, tennis, or squash. Traditional technologies like Hawk-Eye \cite{hawkeye} and View 4D \cite{goalcontrol} often rely on triangulation techniques \cite{triangulation1,triangulation2,triangulation3} for the calculation of the ball's position. A major limitation of these methods is the need for expensive hardware (e.g. multiple synchronized cameras), consequently preventing their application in low budget and amateur sports where only a single camera is used to record the game. In contrast to these methods, a neural network can be trained to predict the 3D position of the ball even if the game is recorded with just a single camera. However, training such a network typically requires 3D ground truth data. Although ground truth data may be readily available for some disciplines such as soccer, it is often lacking in lower-budget sports like squash. Consequently, there is a need for a method that can perform ball localization without relying on 3D ground truth data as supervision. \\[0.75ex]
In this paper, we present a novel method for training a neural network to perform 3D localization that does not need 3D ground truth labels as supervision at all. Instead, only 2D image labels of the position of a desired object are needed, which can be easily obtained through methods such as object detection or manual annotation by clicking on the object's center in an image. Our model is trained using video clips, leveraging the physical laws of motion to enable the network to infer the latent third dimension. \\[0.75ex]
A key advantage of our method is its flexibility in inference. While videos are needed during training, our trained model can also be applied to single images during inference. This ensures that our network can even be applied in situations where the object's motion cannot be precisely described by physics (e.g. due to human interaction) or where the physical parameters are not known completely (e.g. if a sudden wind is affecting the ball). While methods processing multiple frame will likely fail in those situations, our method is robust to such changes. \\[0.75ex]
\begin{figure}[t!]
  \centering    
      \includegraphics[width=0.97\linewidth]{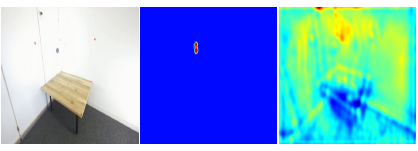}
      \vspace{-0.1cm}
     \caption{Predictions of the \textit{Position Estimation Network}. The left image is the input to the network, the middle image shows the predicted heatmap, and the right image shows the predicted depthmap. By combining the heatmap and depthmap, the 3D position of the ball can be calculated.
     }
     \vspace{-0.1cm}
     \label{img:cover}
     \vspace{-0.50cm}
\end{figure}
\noindent The primary focus of this paper is on training our neural network on video sequences of the ball's movement between contacts with players. During this period, the ball remains unaffected by non-deterministic forces, enabling us to effectively utilize the physical laws of motion to describe its trajectory in the training videos. Importantly, we are able to describe the ball bouncing off walls and the floor, which is e.g. important in various disciplines like tennis or squash. \\[0.75ex]
Since our method is not only limited to sport's applications, we study videos of a moving ball in a more general context. The main contributions of this paper are:
\begin{itemize}[leftmargin=0.5cm,itemsep=0pt, topsep=-0.5\parskip]
  \item We propose a general method to train a neural network for 3D object localization in single images without the need of 3D labels.
  \item We create a synthetic and a real-world dataset with 3D ground truths for the evaluation of our method. The datasets will be published in order to ensure reproducibility and to encourage further research in this field.
  \item We prove experimentally that our model is able to accurately predict the 3D position of the ball and discuss the experimental results thoroughly.
\end{itemize}

%% file: sec/2_relatedwork.tex
\section{Related Work}
\label{sec:relatedwork}

Pose estimation is widely used in many sport disciplines, and it is of great importance in athletics \hbox{\cite{humanpose_athletics}}, table tennis \hbox{\cite{humanpose_tabletennis}}, swimming \hbox{\cite{humanpose_swimming}}, and soccer \hbox{\cite{humanpose_soccer}}. In this paper we estimate the position of an object instead of a human pose. Nevertheless, our architecture is inspired by methods commonly used in human pose estimation, since we use 2D heatmaps to encode the object's position. \\
In addition to human pose estimation, accurate ball localization is crucial in the analysis of many sport disciplines. Several methods have been developed to calculate physical reasonable 3D positions of the ball from the sequence of related 2D positions. For instance, \cite{MonoTrackBadminton} localize a badminton ball using videos, \cite{PhysicsBasedBallTrackingAnd3DTrajectoryReconstructionBasketball} reconstruct the 3D trajectory of a basketball from a sequence of 2D positions and \cite{BallTrackingAnd3DTrajectoryApproximationVolleyball} calculate the trajectory of a volleyball. However, these methods rely on fitting the ball's trajectory to physically calculated paths and, thus, are only applicable to video sequences exhibiting perfect ballistic trajectories. They cannot handle situations where the ball is subject to non-deterministic forces (e.g., human actions) or when the physical parameters change (e.g., the introduction of strong winds or changes in drag coefficients due to wear and tear). Moreover, these methods do not account for ball bouncing. In contrast to these methods, we are also able to describe more intricate physical situations like a ball bouncing off the floor. Furthermore, since our method can be applied to single images during inference, it remains effective even when the ball's behavior is non-deterministic or when precise knowledge of physical parameters is lacking. Thus, our method overcomes many limitations of previous approaches. \\
Other studies like \cite{DeepSportRadarv1,3DBallLocalizationFromASingleCalibratedImage} estimate the 3D position of a basketball in single images by measuring the diameter of the ball in the images. Similarly, \cite{tabletennis} utilizes the ball diameter to predict the 3D position of a table tennis ball. Nevertheless, for these methods additional relatively costly segmentation labels of the ball are needed. In contrast, our method only needs less expensive 2D image-coordinate annotations of the ball either from manual or automatic annotations. \\
To the best of our knowledge, we are the first to develop a method for localizing the ball's 3D center coordinates in single images using only 2D labels together with knowledge of the physical motion. Even though we focus on the 3D localization of a ball, it is worth noting that our method can be applied to other objects as well. \\[0.75ex]
As we utilize the physical knowledge of the moving ball for training our neural network, its physical motion has to be described in a differentiable manner. While the motion can be described by an analytic function for many toy problems, the physical equations of motion have to be solved numerically for most real life applications. We implement the differential equations using the framework of Neural Ordinary Differential Equations (NDEs) \cite{NeuralOrdinaryDifferentialEquations} in order to be able to calculate the gradient of the numerical solution. Our method furthermore resembles Hamiltonian Neural Networks (HNNs) and Lagrangian Neural Networks (LNNs) \cite{HamiltonianNeuralNetworks}. However, in contrast to HNNs and LNNs, we learn the ball's 3D coordinates given a fixed Hamilton function instead of learning the Hamilton or Lagrange function given fixed data.

%% file: sec/3_method.tex
\section{Method}
\label{sec:relatedwork}

\subsection{Overview}

Our method consists of two main modules: A Position Estimation Network (PEN) and a Physics Aware Forecast Module (PAF).
The PEN is a neural network that takes a single image as input and generates a heatmap as well as a depthmap as output. By extracting the 2D image coordinates from the heatmap and using the depth information from the depthmap along with the intrinsic and extrinsic camera calibration matrix, we calculate the camera and world coordinates of the ball.
The PEN is applied to three successive images at time \( t_{n-1} \text{, } t_n \text{, } t_{n+1} \) to obtain the ball's world-coordinates for each image. Using the coordinates at time \( t_{n-1} \) and \( t_{n+1} \), we estimate the ball's velocity at time \( t_n \). 
Given the ball's world coordinates and velocity at time \( t_n \), the PAF calculates the world coordinates at a later time \( t_{n + \Delta n} \) by solving the differential equations of motion. \\
We implement two losses for the training of the PEN: The \textit{heatmap loss} and the \textit{future loss}. The heatmap loss compares the predicted and ground-truth heatmaps to teach the network the ball's 2D image coordinates. The \textit{future loss} ensures that the PEN learns the correct camera depth by projecting the PAF's predicted coordinates at time \( t_{n + \Delta n} \) to the 2D image coordinates for the frame at \( t_{n + \Delta n} \). \\
An overview of our method is depicted in Figure \ref{img:overview}. We note that only the PEN is used during inference.

\subsection{Position Estimation Network (PEN)}
\begin{figure*}[t]
    \centering
    \includegraphics[width=0.8\linewidth]{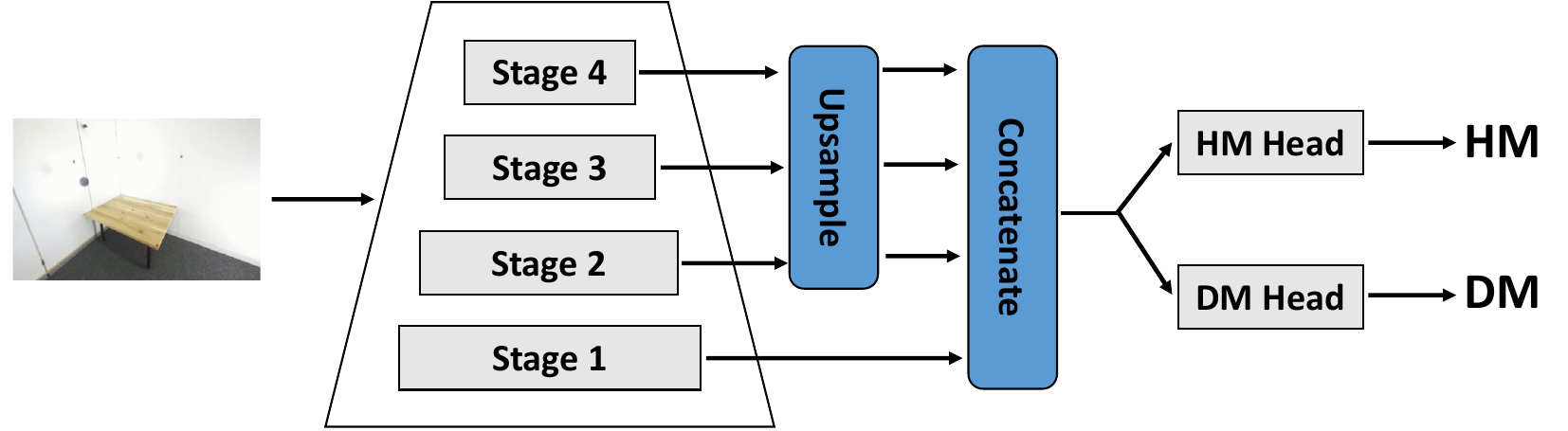}
   \caption{\textit{Position Estimation Network (PEN)} architecture for generating both heatmap (HM) and depthmap (DM) from an input image. Features from the backbone network are concatenated and passed through the two heads. One generates a heatmap and the other produces a depthmap.
   }
   \label{img:sin}
    \vspace{-0.3cm}
\end{figure*}
The PEN takes an image \( I \in \mathbb{R}^{C \times H \times W} \) as input with \( C \), \( H \) and \( W \) being the number of channels, height and width of the image. Using a backbone architecture like ResNet \cite{resnet} or ConvNeXt \cite{convnext, convnextv2}, the features after each stage are bilinearly upsampled to the shape of the first stage and, afterwards, they are concatenated. A \textit{heatmap head} consisting of one \( 1\times 1 \) convolution, two \( 7 \times 7 \) convolutions and a final \( 1\times 1 \) convolution is applied to obtain a heatmap \( \Tilde{H} \in \mathbb{R}^{H \times W} \) from the concatenated features. Additionally, a separate \textit{depthmap head} with the same architecture is applied to the concatenated features to calculate the depthmap \( D \in \mathbb{R}^{H \times W} \). We intentionally choose a simple architecture for the heads to demonstrate the effectiveness of our method without requiring extensive architecture modifications. The architecture of the PEN is depicted in Figure \ref{img:sin}. \\[0.75ex]
The 2D image coordinates \( \va{r}^{(\mathrm{I})} = \begin{pmatrix} x^{(\mathrm{I})} & y^{(\mathrm{I})} \end{pmatrix}^\mathrm{T} \) are extracted from the heatmap with a 2D soft-argmax by weighting each pixel-position with its corresponding heatmap value. Furthermore, the camera depth \( z^{(\mathrm{C})} \) is calculated as the sum over all values of the depthmap weighted with the corresponding values of the 2D softmax of the heatmap. Consequently, the coordinates are calculated from the heatmap \( \Tilde{H} \) and the depthmap \( \Tilde{D} \) as
\begin{align}
    \label{eq:cameradepth}
    \begin{split}
    x^{(\mathrm{I})} = \sum\limits_{h = 0}^{H-1} \sum\limits_{w = 0}^{W-1}  w \cdot \left( \text{softmax}\left(\beta \cdot \Tilde{H} \right) \right)_{h,w}
    \\
    y^{(\mathrm{I})} = \sum\limits_{h = 0}^{H-1} \sum\limits_{w = 0}^{W-1}  h \cdot \left( \text{softmax}\left(\beta \cdot \Tilde{H} \right) \right)_{h,w}
    \\
    z^{(\mathrm{C})} = \sum\limits_{h = 0}^{H-1} \sum\limits_{w = 0}^{W-1} \Tilde{D}_{h,w} \cdot \left( \text{softmax}\left(\beta \cdot \Tilde{H} \right) \right)_{h,w}
    \end{split}
\end{align}
with the constant factor \( \beta = 10 \) to obtain sharp softmax probabilities and the pixel indices \( h \) and \( w \). 

\subsection{Coordinate Transformations}
We assume that the intrinsic camera matrix \( \textbf{M}_\text{int} \) and the extrinsic camera matrix \( \textbf{M}_\text{ext} \) are known, and we discuss in the supplementary material in more detail how the camera matrices can be obtained. \\
We use homogeneous coordinates and transform world coordinates into camera coordinates with the extrinsic matrix \( \textbf{M}_\text{ext} \in \mathbb{R}^{4\times 4} \) and  camera coordinates into image coordinates with the intrinsic matrix \( \textbf{M}_\text{int} \in \mathbb{R}^{3\times 3} \). Since these transformations are invertible we calculate the camera-coordinates \( \va{r}^{(\mathrm{C})} = \begin{pmatrix} x^{(\mathrm{C})} & y^{(\mathrm{C})} & z^{(\mathrm{C})} \end{pmatrix}^\mathrm{T} \) and the world-coordinates \( \va{r}^{(\mathrm{W})} = \begin{pmatrix} x^{(\mathrm{W})} & y^{(\mathrm{W})} & z^{(\mathrm{W})} \end{pmatrix}^\mathrm{T} \) given the image coordinates \( \va{r^{(\mathrm{I})}} \) as well as the camera depth \( z^{(\mathrm{C})} \) using the following formulas:
\begin{align}
  \resizebox{.90\linewidth}{!}{$
  \begin{pmatrix}
    x^{(\mathrm{C})} \\
    y^{(\mathrm{C})} \\
    z^{(\mathrm{C})}
  \end{pmatrix}
  = \textbf{M}_\text{int}^{-1} \cdot 
  \begin{pmatrix}
    x^{(\mathrm{I})} z^{(\mathrm{C})} \\
    y^{(\mathrm{I})} z^{(\mathrm{C})} \\
    z^{(\mathrm{C})}
  \end{pmatrix} 
  \hspace{0.1cm} \text{,} \hspace{0.25cm}
  \begin{pmatrix}
    x^{(\mathrm{W})} \\
    y^{(\mathrm{W})} \\
    z^{(\mathrm{W})} \\
    1
  \end{pmatrix}
  = \textbf{M}_\text{ext}^{-1} \cdot 
  \begin{pmatrix}
    x^{(\mathrm{C})} \\
    y^{(\mathrm{C})} \\
    z^{(\mathrm{C})} \\
    1
  \end{pmatrix} .
$}
\end{align}
Furthermore, the velocity \( \va{v}(t_n) \) at time \( t_n \) is calculated from the ball's world coordinates at time \( t_{n-1} \) and \( t_{n+1} \) using the symmetric finite difference quotient as
\begin{align}
    \label{eq:velocity}
    \va{v}(t_n) = \frac{\va{r}^{(\mathrm{W})}(t_{n+1}) - \va{r}^{(\mathrm{W})}(t_{n-1})}{t_{n+1} - t_{n-1}} .
\end{align}

\subsection{Physics Aware Forecasting Module (PAF)}
Given the ball's world coordinates \( \va{r}^{(\mathrm{W})}(t_n) \) and velocity \( \va{v}^{(\mathrm{W})}(t_n) \) at time \( t_n \) as initial conditions, the PAF calculates the ball's coordinates at a later time \( t_{n+ \Delta n} \) by solving the differential equations of motion. While the ball's motion can be expressed as an analytical function for many simple toy problems, it is not always possible to find an analytical solution for more intricate situations. Since we want to describe a bouncing ball in an arbitrary environment, we solve the differential equations of motion numerically. By using the framework of Neural Differential Equations \cite{NeuralOrdinaryDifferentialEquations} we are able to calculate the gradient of the numerical solution that is needed for the backpropagation algorithm. \\[0.75ex]
We adopt the Hamilton formalism to describe the motion of the ball. A physical system can be described according to \cite{hamiltonmechanics} using the Hamilton function
\begin{align}
  \label{eq:hamiltonian}
      \mathcal{H} = \frac{\left| \va{p} \right|^2}{2m} + V(\va{r}^{(\mathrm{W})})
\end{align}
with the ball's mass \( m \), the momentum \( \va{p} = m \va{v} \) (for euclidean coordinates) and the potential \( V(\va{r}^{(\mathrm{W})}) \), which describes the physical behavior of the ball. According to the Hamilton formalism \cite{hamiltonmechanics}, the equations of motion are derived from the Hamilton function as 
\begin{align}
  \label{eq:hamiltonequations}
  \frac{\mathrm{d}}{\mathrm{dt}} \va{r}^{(\mathrm{W})} = \frac{1}{m}\frac{\mathrm{d}}{\mathrm{d}\va{v}} \mathcal{H}
  \hspace{0.2cm} \text{, } \hspace{0.7cm}
  m\frac{\mathrm{d}}{\mathrm{dt}} \va{v} = - \frac{\mathrm{d}}{\mathrm{d}\va{r}^{(\mathrm{W})}} \mathcal{H}
\end{align}
and by solving these differential equations given the initial ball position and velocity we get the ball's position \( \va{r}^{(\mathrm{W})}(t) \) at time \( t \). Consequently, the equations \ref{eq:hamiltonequations} are solved in the PAF to obtain the ball's position at a later time. \\[0.75ex]
The potential \( V(\va{r^{(\mathrm{W})}}) \) defines the physical behavior of the ball and, thus, we have to model it accordingly to ensure the correct prediction of the ball's movement. The potential of a ball in free fall above the floor at \( z^{(\mathrm{W})} = 0 \) is described as 
\begin{align}
  \label{eq:potential_g}
  V_\mathrm{G}( \va{r}^{(\mathrm{W})}) = m \cdot g \cdot \text{ReLU}(z^{(\mathrm{W})})
\end{align}
with \( g \) being the gravitational constant.
Additionally, we also want to describe the bouncing off the floor. Ideally, the potential at the floor position is described as an infinite potential barrier. However, it is sufficient to approximate the infinite barrier as a fast increasing potential. Therefore, we model the potential at the floor position as 
\begin{align}
  \label{eq:potential_f}
  V_\mathrm{F}( \va{r}^{(\mathrm{W})}) = m \cdot c \cdot \text{ReLU}(-z^{(\mathrm{W})})
\end{align}
with \( c = \SI{1000}{\frac{J}{m\cdot kg}} \) being a large constant. We note that the exact value of \( c \) is not important as long as it is significantly larger than \( g \). \\[0.75ex]
In some environments, we also have to include the bouncing off the walls or various obstacles. These can be defined similarly and are explained in the supplementary material for our specific datasets in more detail. As a result, the overall potential of a moving ball is described as
\begin{align}
  \label{eq:potential}
  \resizebox{.90\linewidth}{!}{$
            \displaystyle
  V( \va{r}^{(\mathrm{W})}) = V_\mathrm{G}( \va{r}^{(\mathrm{W})}) + V_\mathrm{F}( \va{r}^{(\mathrm{W})}) + V_\mathrm{W}( \va{r}^{(\mathrm{W})}) + V_\mathrm{O}( \va{r}^{(\mathrm{W})})
  $}
\end{align}
with \( V_\mathrm{W}( \va{r}^{(\mathrm{W})}) \) describing the bouncing off the walls and \( V_\mathrm{O}( \va{r}^{(\mathrm{W})}) \) describing the bouncing off obstacles. \\[0.75ex]
After defining the potential for a specific physical situation, we implement the differential equation \ref{eq:hamiltonequations} in the PAF. In our experiments, we use the 5\textsuperscript{th} order Dormand-Prince method \cite{dopri5} to solve the differential equations numerically, but we note that other solvers can be used as well.

\begin{figure*}[t]
  \centering
  \begin{subfigure}[l]{0.235\linewidth}
    \includegraphics[width=\textwidth]{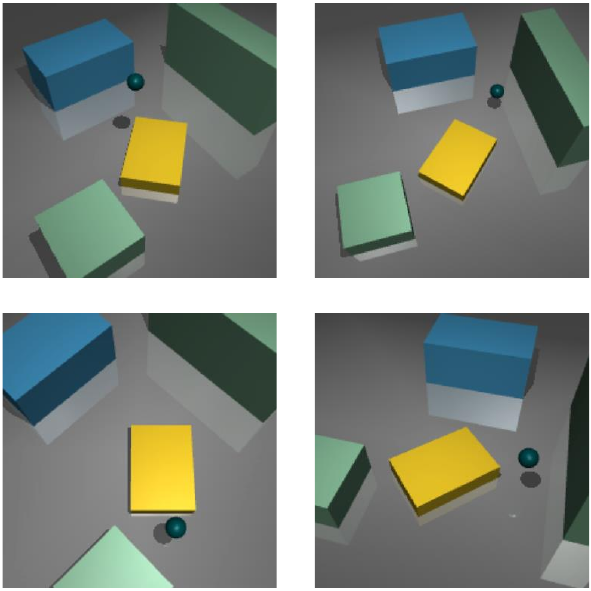}
    \caption{\textit{SD}: images captured from different camera locations in the 1\textsuperscript{st} environment.}
    \label{img:dataset_cams}
  \end{subfigure}
  \hspace{2cm}
  \begin{subfigure}[r]{0.38\linewidth}
    \begin{subfigure}[b]{\textwidth}
      \includegraphics[width=\textwidth]{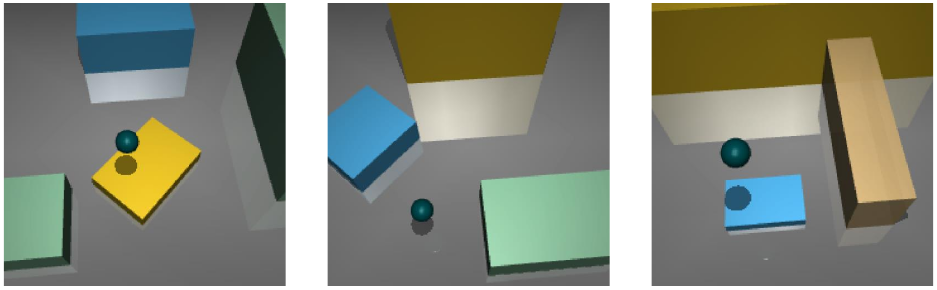}
      \caption{\textit{SD}: images captured in different environments.}
      \label{img:dataset_envs}
    \end{subfigure}
    \vfill
    \begin{subfigure}[b]{\textwidth}
      \includegraphics[width=\textwidth]{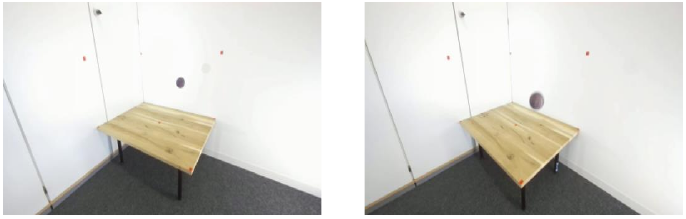}
      \caption{\textit{RD}: images captured from different camera locations.}
      \label{img:dataset_real}
    \end{subfigure}
  \end{subfigure}
  \vspace{-0.2cm}
  \caption{Example images from the \textit{synthetic dataset} (\textit{SD}) and \textit{real dataset} (\textit{RD}).}
  \vspace{-0.2cm}
  \label{img:dataset_example}
\end{figure*}

\subsection{Loss}
Since we do not use 3D ground truth labels, we utilize the \textit{heatmap loss} and the \textit{future loss} to train the PEN. As it is common in human pose estimation (e.g. \cite{TokenPose}), we also use heatmaps to teach the model to predict the 2D image coordinates. We calculate the \( \mathrm{L}_2 \) loss between the heatmap output \( \Tilde{H}(t_n) \) of the PEN for the frame at time \( t_n \) and the ground truth heatmap \( \Tilde{H}_\text{gt}(t_n) \). This ground truth heatmap is a two-dimensional Gaussian centered at the ground truth image coordinates at time \( t_n \). \\
The \textit{future loss} utilizes the prediction of the PAF to teach the model the correct prediction of the camera depth. First, we transform the prediction of the PAF at time \( t_{n + \Delta n} \) to image coordinates using the camera matrices and obtain the image coordinates \( \va{r}^{(\mathrm{I})}_\text{PAF} \). Then, we calculate the \( \mathrm{L}_1 \) loss between these coordinates and the image coordinates calculated by the PEN for the frame at time $t_{n + \Delta n}$. \\[0.75ex]
Consequently, the total loss is calculated as 
\begin{equation}
  \label{eq:loss}
  \begin{split}
  \mathrm{L} &= 
  {\| \va{r}^{(\mathrm{I})}_\text{PEN}(t_{n + \Delta n}) - \va{r}^{(\mathrm{I})}_\text{PAF}(t_{n + \Delta n}) \|}_\text{L1} 
  \\
  &+ \frac{1}{\mathcal{ \abs{T} }} \sum\limits_{t_i \in \mathcal{T}} {\| \Tilde{H}(t_i) - \Tilde{H}_\text{gt}(t_i) \|}_\text{L2}
  \end{split}
\end{equation}
with \( \mathcal{T} = \{ t_{n-1}, t_{n}, t_{n+1}, t_{n + \Delta n} \} \).

\subsection{Evaluation Metrics}
In order to test the quality of the PEN's 3D predictions we introduce the \textit{distance to groundtruth (DtG)} metric. It measures the euclidian distance between the prediction and the ground truth 3D position. Accordingly, the metric is calculated as
\begin{equation}
  \label{eq:DtG}
  \textit{DtG} = \frac{1}{\left| \mathcal{V} \right|} \sum\limits_{t_i \in \mathcal{V} } \norm{ \va{r}_\text{gt}^{(\mathrm{C})}(t_i) - \va{r}_\text{PEN}^{(\mathrm{C})}(t_i) }_\text{L2}
\end{equation}
over all images in the test set $\mathcal{V}$. We note that the metric remains the same regardless of whether world coordinates or camera coordinates are used. \\
Since we expect the estimation of the ball's 3D position to be less accurate if the ball is further away from the camera, we define the \textit{binned distance to groundtruth (DtG\(_b\))} metric. Depending on the ball's groundtruth camera depth \( z^{(\mathrm{C})}_\text{gt} \), the predictions are grouped into \( b \) bins and the \textit{DtG} is calculated for each bin separately. Thus, it is calculated as
\begingroup
\renewcommand*{\arraystretch}{1.05}
\begin{align}
  \label{eq:DtGb}
  \resizebox{.83\linewidth}{!}{$
  \textit{DtG}_b = \begin{pmatrix}
    \frac{1}{\left| \mathcal{V}_0 \right|} \sum\limits_{t_i \in \mathcal{V}_0 } \norm{ \va{r}_\text{gt}^{(\mathrm{C})}(t_i) - \va{r}_\text{PEN}^{(\mathrm{C})}(t_i) }_\text{L2} \\[-0.65em]
    \vdotswithin{\frac{1}{\left| \mathcal{V} \right|} \sum\limits_{t_i \in \mathcal{V} } \norm{ \va{r}_\text{gt}^{(\mathrm{C})}(t_i) - \va{r}_\text{PEN}^{(\mathrm{C})}(t_i) }_\text{L2}} \\[0.3em]
    \frac{1}{\left| \mathcal{V}_{b-1} \right|} \sum\limits_{t_i \in \mathcal{V}_{b-1} } \norm{ \va{r}_\text{gt}^{(\mathrm{C})}(t_i) - \va{r}_\text{PEN}^{(\mathrm{C})}(t_i) }_\text{L2}
  \end{pmatrix}
  $}
\end{align}
\endgroup
with \( \mathcal{V}_j = \{ t_i \in \mathcal{V} \mid z^{(\mathrm{C})}_\text{gt}(t_i) \in [ j \cdot \frac{z_\text{max} - z_\text{min}}{b} + z_\text{min}, (j + 1) \cdot \frac{z_\text{max} - z_\text{min}}{b} + z_\text{min} ] \} \) and \( j \in \{ 0, \text{...}, b-1 \} \). We select suitable values for the number of bins \( b \), the minimal distance \( z_\text{min} \) and the maximal distance \( z_\text{max} \) depending on each dataset. 

%% file: sec/4_dataset.tex
\section{Dataset}
\label{sec:dataset}

\label{sec:datasets}
To test our method we create two datasets consisting of monocular videos of a moving ball: the \textit{synthetic dataset (SD)} and the \textit{real dataset (RD)}. For the training we provide the ball's 2D positions for every frame and for the evaluation we also provide the ball's 3D position. Some example images are depicted in Figure \ref{img:dataset_example} and more images are visualized in the supplementary material.

\subsection{Synthetic Dataset}
We create the \textit{SD} using the general purpose physics engine MuJoCo \cite{MuJoCo} and, thus, we are able to create synthetic data with realistic physical behavior. We simulate a moving ball under the influence of gravity. The ball bounces off an infinite floor and additional finite obstacles placed in the environment. We generate 3 different environments and for each environment we capture the scenes from 9 different camera location. However, each distinct scene is only captured from a single camera in contrast to traditional triangulation and binocular vision settings. The cameras are placed such that the distance to the origin of the scene is between \( \SI{8}{m} \) and \( \SI{10}{m} \), and the exact camera locations are given in the supplementary material. The videos are recorded with \SI{30}{\fps}. We create 3 versions of the dataset:
\begin{itemize}[leftmargin=0.5cm,itemsep=0pt, topsep=-0.5\parskip]
  \item \textit{SD-S}: We use a single environment and capture the scene from a single fixed camera location. We create 100 videos for the training set, 100 for the validation set and 100 for the test set.
  \item \textit{SD-M}: We use a single environment and capture the scene from 9 different camera locations. The training set consists of 100 videos per camera location recorded from camera locations 1-6, while the validation and test set each contain 100 videos per camera location recorded from all 9 camera locations.
  \item \textit{SD-L}: We include videos recorded in 3 different environments and for each environment the scene can be captured from 9 different camera locations. For each environment, the training set contains 100 videos per camera location recorded from camera locations 1-6, while the validation and test set contain 100 videos per camera location recorded from all 9 camera locations.
\end{itemize}
The versions are designed such that \textit{SD-M} is a subset of \textit{SD-L} and \textit{SD-S} is a subset of \textit{SD-M}. Each image has a resolution of \( 224 \times 224 \) pixel and each clip consists of 30 consecutive frames during training. Example frames for different camera locations in the 1\textsuperscript{st} environment are shown in Figure \ref{img:dataset_cams}, and for the three different environments in Figure \ref{img:dataset_envs}. Random initial positions and velocities of the ball are selected for each video to ensure diversity among the samples. \\
By applying our method to \textit{SD-S} we are able to show that our model is capable of learning the ball's 3D position in general. However, real use cases like sport broadcasting might not use a fixed camera, instead the camera sometimes moves and changes its location between the scenes. Therefore, we use the \textit{SD-M} to show that our model is also able to learn the ball's 3D position if the camera changes its location. Finally, it is beneficial for practical applications to train a model that is working on different environments like different known courts in sport broadcasting. Using the \textit{SD-L}, we show that our model is also able to learn the ball's 3D position even if the environment changes.
By applying our method to these three synthetic datasets we test some characteristics needed for practical application.

\subsection{Real Dataset}
Since the synthetic images are generated by a physics engine, they are visually not perfectly realistic. Therefore, we create the \textit{RD} to prove that our method is also able to cope with noisy input data. We use a ZED 2i stereo camera \cite{zed} to record a rubber ball bouncing off the floor, the walls and an additional obstacle. We record each video from one of two distinct camera locations and use a frame rate of \SI{60}{\fps}. The distance of the camera to the origin of the scene is between \( \SI{1.3}{m} \) and \( \SI{1.4}{m} \). Only the data of the left camera is used as input to the PEN, and we manually labelled the 2D image coordinates for each frame. We depict example frames from the two different camera locations in Figure \ref{img:dataset_real}. Each frame is resized to \( 224 \times 384 \) pixel. We split the videos into clips consisting of 16 frames each, with 251 clips in the train set, 34 clips in the validation set, and 81 clips in the test set. In order to evaluate our method we use the data of the stereo camera to calculate a ground truth depthmap of the scene. The ground truth camera depth is obtained from the depthmap value at the 2D image coordinates of the ball. By applying our method to this dataset we are able to show that it also works with simple yet realistic input data.

%% file: sec/5_experiments.tex
\vspace{-0.1cm}
\section{Experiments}
\label{sec:experiments}

In this section we perform multiple experiments and apply our method to the datasets described in section \ref{sec:datasets}. We use the first 4 stages of a ResNet34 \cite{resnet} pretrained on ImageNet \cite{imagenet} as the backbone of the PEN and provide additional experiments with different backbones in the supplementary material. To ensure performance stability, we maintain a second copy of our model whose weights are an exponential moving average (EMA) of the trained model weights, and we update the EMA model after each iteration. At the end of each epoch the EMA model is evaluated on the validation set, and we only keep the model with the best \textit{DtG} score. We calculate the \textit{future loss} at multiple time steps. For the synthetic data we use $\Delta n = 4$, $\Delta n = 8$ and $\Delta n = 15$ and for the real data we use $\Delta n = 2$, $\Delta n = 4$ and $\Delta n = 6$, because the average speed of the real ball is larger than the speed of the synthetic ball. Our code and datasets will be published at \hbox{www.example.com} where we will also provide additional visual evaluations of our experiments.

\subsection{Synthetic Experiments}
\begin{figure*}[t]
  \centering
  \begin{minipage}{0.71\linewidth}
    \centering
    \includegraphics[width=\linewidth]{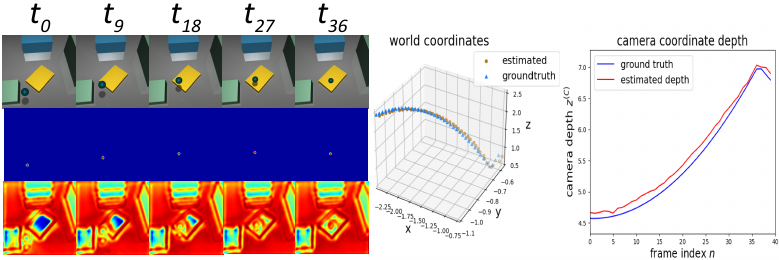}

    \captionof{figure}{Predictions of the \textit{SD}-L model on a sample video from the \( 1^\text{st} \) camera location in the \( 1^\text{st} \) environment. The first row shows the input images, the second row shows the predicted heatmaps, and the third row shows the predicted depthmaps. On the right, a 3D plot of the predicted ball trajectory \(\va{r}^{(\mathrm{W})} \) and a 2D plot of the camera depth \(z^{(\mathrm{C})} \) is shown.
    }
    \label{img:results2}
  \end{minipage}
  \hfill
  \begin{minipage}{0.22\linewidth}
    \centering
    \includegraphics[width=\linewidth]{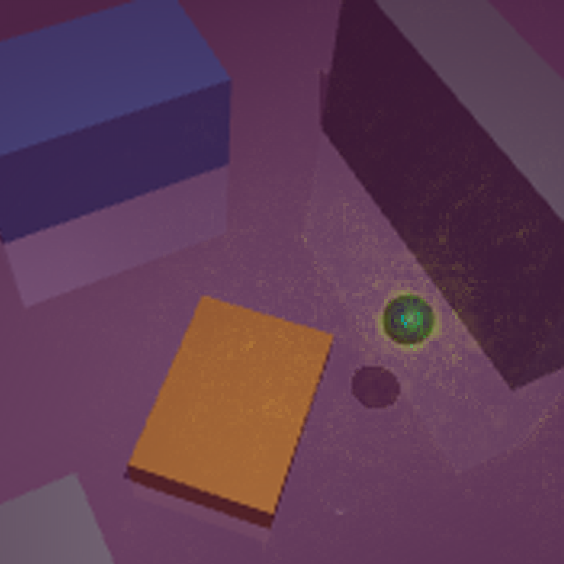}
    \captionof{figure}{Saliency Map for a prediction of the SD-L model.}
    \label{img:saliency_sd}
  \end{minipage}
  \vspace{-0.3cm}
\end{figure*}
In this section we evaluate our method on the synthetic datasets described in section \ref{sec:datasets}. We train three models, one for each subset of the synthetic datasets:
\begin{itemize}[leftmargin=0.5cm,itemsep=0pt, topsep=-0.5\parskip]
  \item \textit{SD-S}: The first model is trained solely on the \textit{SD-S} training set. Therefore, the model is only exposed to images from a single camera location and a single environment during both training and testing.
  \item \textit{SD-M}: The second model is trained on the \textit{SD-M} training set, which includes images from the first 6 camera locations of the 1\textsuperscript{st} environment. However, during evaluation, it is tested on videos recorded from all camera locations of the 1\textsuperscript{st} environment. By comparing its results on camera 1 to the results of the first model, we test if training on multiple camera locations improves the model's performance. By looking at the results on the last three camera locations we test if the model is able to generalize to previously unseen camera locations.
  \item \textit{SD-L}: The third model is trained on the \textit{SD-L} training set, which includes images from the first 6 camera locations of each environment. During evaluation, it is tested on all camera locations of each environment in the test set. With this model we are able to test whether the model is able to learn the ball's position for different known environments. Moreover, we test if the model benefits from being trained on multiple environments by comparing the results of this model to the results of the second model on the 1\textsuperscript{st} environment.
\end{itemize}
The results on the test set are given in Table \ref{tab:DtG_synthetic}. For the three models, we evaluate the average \textit{DtG} sores on the videos from exemplary camera locations in the 1\textsuperscript{st} environment in Table \ref{tab:DtG_synthetic_cam}. Since the performance of the \textit{SD-M} model is significantly better than the performance of the \textit{SD-S} model, we conclude that the PEN is not just able to deal with input from multiple camera perspectives, but it even benefits from being trained with additional data from multiple camera location. Moreover, we see that the model is able to generalize to previously unseen camera locations since the scores for camera 7 to 9 are also good. By comparing the \textit{SD-L} model to the \textit{SD-M} model, we see that the \textit{SD-L} model achieves even slightly better scores and, consequently, we conclude that it is possible to apply the PEN to multiple known environments. In Table \ref{tab:DtG_synthetic_env} we see the average \textit{DtG} scores on the three environments. The \textit{SD-L} model performs again better than the \textit{SD-M} model, thus, reinforcing the impression that training the model on multiple environments is beneficial. The results of the \textit{SD-L} model are visualized in Figure \ref{img:results2}, and it is visible that the predictions match the 3D ground truth very accurately.\\[0.75ex]
To conclude, we see that it is not only possible to train our model with distinct videos from multiple camera locations, but even recognize the fact that using distinct videos from multiple camera locations is beneficial. This is important for the application to realistic data (e.g. sports broadcasting) as in many cases the camera is changing its position instead of being stationary. Moreover, we also show that our method is capable of dealing with multiple known environments, which is also an important property for practical applications (e.g. different stadiums or courts in sport broadcasting).

\begin{figure*}[t]
  \centering
  \begin{minipage}{0.75\linewidth}
    \centering
    \includegraphics[width=\linewidth]{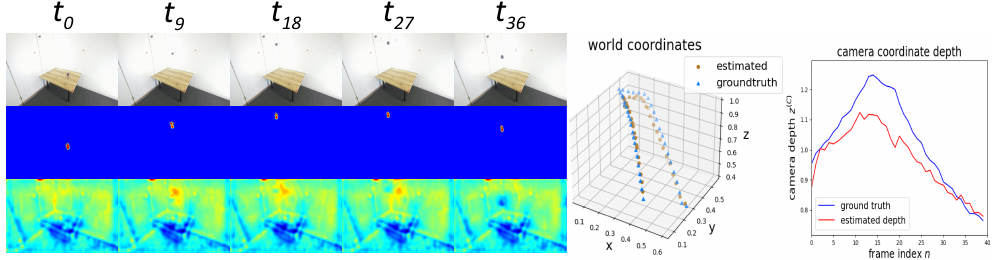}
    \captionof{figure}{PEN's predictions on a sequence from the \textit{real dataset}. The first row shows the input images, the second row shows the predicted heatmaps, and the third row shows the predicted depth map. On the right, a 3D plot of the predicted ball trajectory \(\va{r}^{(\mathrm{W})} \)  and the camera depth \(z^{(\mathrm{C})} \) is shown.
    }
    \label{img:results}
  \end{minipage}
  \hfill
  \begin{minipage}{0.22\linewidth}
    \centering
    \includegraphics[width=0.9\linewidth]{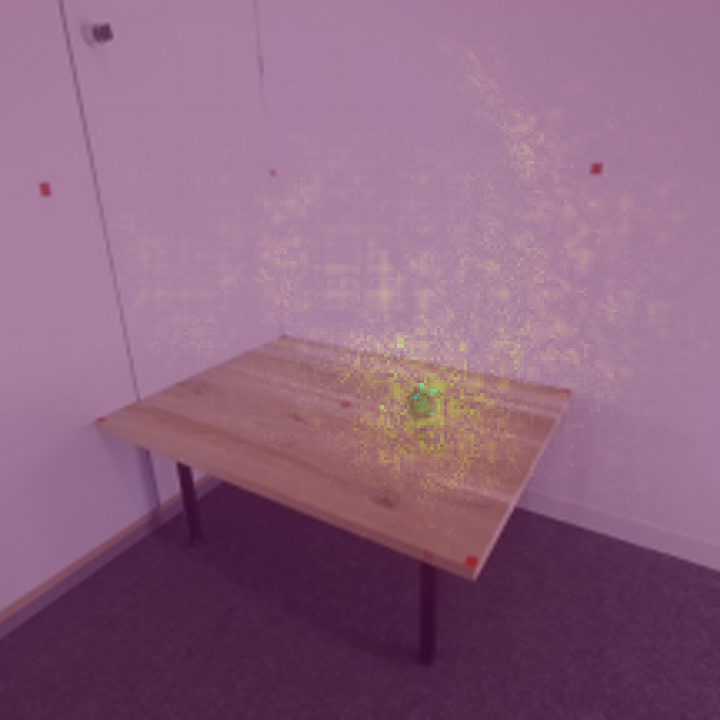}
    \captionof{figure}{Saliency map for a prediction of the RD model.}
    \label{img:saliency_rd}
  \end{minipage}
  \vspace{-0.2cm}
\end{figure*}

\subsection{Real Experiments}
Even though we were able to experimentally prove some properties needed for practical applications on the \textit{synthetic dataset}, we still need to show that our model is able to work with real data. Important difficulties in working with real videos are that the model has to be able to filter out pixel noise in the image data and that it needs to be capable of working with noisy labels. Because the 2D image labels are manually annotated and do not always perfectly match the center of the ball, additional noise is also added to the calculation of the future loss. Furthermore, the physical description of the ball might not perfectly match the ball's real behavior (e.g. due to deformations of the ball or spin), resulting in reduced precision of the forecasting. To experimentally demonstrate that our model is able to cope with these difficulties, we evaluate our method on the \textit{real dataset}.
\begin{table}[h]
  \centering
  \captionsetup{width=\textwidth}
  \small
  \caption{\small Scores per camera location on the \textit{real dataset}. \normalsize}
  \vspace{-0.1cm}
  \label{tab:DtG_real}
  \resizebox{0.9\linewidth}{!}{
  \begin{tabular}{ccccc}
    \toprule
     & \multicolumn{2}{c}{ \( \textit{DtG} \pm \Delta \textit{DtG} \) (cm) \(\downarrow\)} & \multicolumn{2}{c}{ \( \textit{DtG}_3 \pm \Delta \textit{DtG}_3 \) (cm) \(\downarrow\)} \\
    \cmidrule(r){2-5}
    training set & camera 1 & camera 2 & camera 1 & camera 2\\
    \midrule
    \textit{RD} & 
    \( 7 \pm 4 \) & 
    \( 6 \pm 4 \) & \( 
      \begin{pmatrix}
        8 \pm 4 \\
        6 \pm 2 \\
        11 \pm 4
      \end{pmatrix} \) & \( 
        \begin{pmatrix}
          4 \pm 2 \\
          6 \pm 3 \\
          11 \pm 4
        \end{pmatrix} \) \\
    \bottomrule
  \end{tabular}
  }
  \normalsize
  \vspace{-0.2cm}
\end{table}
The results on the test set of the real dataset are shown in Table \ref{tab:DtG_real}. We achieve a \textit{DtG} score of \( \SI{7}{cm} \) for the images recorded from the 1\textsuperscript{st} camera location and a \textit{DtG} score of \( \SI{6}{cm} \) for the images from the 2\textsuperscript{nd} camera location, thus achieving very precise predictions. Consequently, our method is able to deal with the difficulties of real data. \\
Furthermore, we calculate the \( \textit{DtG}_3 \) scores by dividing the data into 3 bins according to equation \ref{eq:DtGb} with \( z_\text{min} = \SI{5}{cm} \) and \( z_\text{max} = \SI{2}{m} \). By analyzing the individual bins, we see that the scores in the 3\textsuperscript{rd} bin are noticeably worse. This can be explained by the fact that it is harder for the model to predict the ball's location when it is farther away from the camera. However, the predictions are still very accurate in the third bin. Nevertheless, further research on this scaling behavior is needed for the next step towards future applications. In figure \ref{img:results} we show the predictions of the model for an example video sequence. Looking at the 3D plot, it is also clear that the predicted trajectory of the ball matches the ground truth trajectory very closely.

\subsection{Emergence of Depthmaps}
Figure \ref{img:results2} and \ref{img:results} clearly reveal the emergence of the scene's geometric structure in the depthmaps, despite the fact that precise depth prediction is only necessary in the vicinity of the ball to produce accurate outputs. Thus, it appears that the model is able to learn a full depthmap of the environment. This observation highlights an intriguing characteristic of our method that raises the possibility of indirect training for monocular depth estimation. However, a comprehensive analysis of this phenomenon is beyond the scope of this paper, and future research is necessary to evaluate this aspect in more detail.

\subsection{Interpretation of the learning process}
In this section, we delve into the question of what information the PEN utilizes in the images to calculate the depth. It is possible that the network simply measures the diameter of the ball in the image for the calculation of the depth (similar to \cite{3DBallLocalizationFromASingleCalibratedImage}). Another possibility is that the network compares the ball to its surrounding environment.  \\
To gain a rough understanding of the regions in the image that are important for the calculation of the depth, we calculate a saliency map \( S \) for the input images as
\begin{equation}
  \resizebox{0.30\linewidth}{!}{$
  S_{i, j} = \sum\limits_{c=0}^{3} \frac{\partial z^{(\mathrm{C})}}{\partial I_{c, i, j}}
  $}
\end{equation}
where \( I \) is the input image, \( z^{(\mathrm{C})} \) is the predicted depth of the ball, \( i \) and \( j \) index the pixels of the image, and \( c \) denotes the channel index. In Figure \ref{img:saliency_sd} and \ref{img:saliency_rd} we depict the saliency map for a real as well as a synthetic image. We observe the highest values around the ball, indicating that the PEN focuses mainly on the size of the ball for the estimation of the depth. However, we also see that the PEN considers the local surroundings of the ball, suggesting that this additional information enhances the depth estimation beyond a simple diameter measurement. Furthermore, we conjecture that the learned depthmap of the scene is a result of the PEN attending to the local context around the ball.

%% file: sec/6_conclusion.tex
\section{Conclusion}
\label{sec:conclusion}

This paper has introduced a novel approach for monocular 3D object localization and demonstrated its effectiveness through experiments on both synthetic and real datasets. Our proposed method eliminates the need for 3D labels as supervision by leveraging the physical equations of motion, allowing the model to infer the latent third dimension. We discussed various properties relevant to real-world scenarios and demonstrated the robustness of our method in handling them. Consequently, our method is a promising step towards 3D object localization applications, although further research is needed to enhance its practical usability. Therefore, we intend to conduct further research to refine and expand upon our method, focusing on areas like scalability, interpretability, and generalization ability.

%% file: sec/X_suppl.tex
\clearpage
\setcounter{page}{1}
\maketitlesupplementary

\begin{abstract}
    In this supplementary material, we provide additional details on the methodology, offer further descriptions and visualizations of the datasets, and present additional results. The code and dataset is published at \url{https://kiedani.github.io/3DV2024/}.
\end{abstract}

\section{Methods}

\subsection{Derivation of the Differential Equations of Motion}
In order to define the differential equations of motion \ref{eq:hamiltonequations}, we first need to model the full potential \ref{eq:potential}. While the gravitational potential is already defined in equation \ref{eq:potential_g} and the bouncing off the floor is modeled through equation \ref{eq:potential_f}, we still need to define \( V_\mathrm{W} \) and \( V_\mathrm{O} \). The potential \( V_\mathrm{W} \) models the interaction of the ball with infinite walls of the room and, thus, it is very similar to \( V_\mathrm{f} \). Since there are no walls in the synthetic dataset, we set \( V_\mathrm{W} = 0 \). In the real dataset are two walls, one in the \(y \text{-} z \) plane at \( x = \SI{0}{m} \) and one in the \( x \text{-} z \) plane at \( y = \SI{0.6}{m} \). Thus, the ball is constrained by the walls and the floor to the area \( x \ge \SI{0}{m} \), \( y \le \SI{0.6}{m} \) and \( z \ge \SI{0}{m} \). Therefore, we model the potential \( V_\mathrm{W} \) as
\begin{align}
    \resizebox{0.93\linewidth}{!}{$
  V_\mathrm{W}( \va{r}^{(\mathrm{W})}) = m c \cdot \text{ReLU}(-x^{(\mathrm{W})}) + m c \cdot \text{ReLU}(y^{(\mathrm{W})} - \SI{0.6}{m} ) 
  $}
\end{align}
with \( c = 1000 \) being a large constant. \\[0.75ex]
The collisions with finite obstacles are described by the potential \( V_\mathrm{O} \), and there are multiple obstacles in the three different environments of the synthetic dataset and one obstacle in the real dataset. To model the potential, we consider a cubic object lying on the floor at the origin of the so-called \textit{object coordinate system} with the length \( 2l \), width \( 2w \) and height \( h \) such that the \textit{object coordinates} $\va{r}^{(\mathrm{o})} = \begin{pmatrix} x^{(\mathrm{o})} & y^{(\mathrm{o})} & z^{(\mathrm{o})}\end{pmatrix}^\mathrm{T}$ with \hbox{$x^{(\mathrm{o})}\in [-l\text{, }l] \land y^{(\mathrm{o})}\in [-w\text{, }w] \land z^{(\mathrm{o})}\in [0\text{, }h]$} describe points inside the cube. To accurately model collisions with this object, the potential has to be large around the boundaries of the cube and zero everywhere outside the cube. In this situation, the potential \( V_o \) describing collisions with this single obstacle at the origin can be described as
\begin{align}
    \small
  \begin{split}
  V_{o}(\va{r}^{(\mathrm{o})}) &= m \frac{c}{\beta_\mathrm{o}} \cdot \left( \sigma \left(\beta_\mathrm{o} (x^{(\mathrm{o})} + l) \right) - \sigma \left(\beta_\mathrm{o} (x^{(\mathrm{o})} - l) \right) \right) \\
    &\cdot \left( \sigma \left(\beta_\mathrm{o} (y^{(\mathrm{o})} + w) \right) - \sigma \left(\beta_\mathrm{o} \cdot (y^{(\mathrm{o})} - w) \right) \right) \\
    &\cdot \left( \sigma \left(\beta_\mathrm{o} \cdot (z^{(\mathrm{o})}) \right) - \sigma \left(\beta_\mathrm{o} \cdot (z^{(\mathrm{o})} - h) \right) \right)
\end{split}
\normalsize
\end{align}
using the sigmoid function \( \sigma (\cdot ) \) and the large constant \( \beta_\mathrm{o} = 66 \) to sharpen the softmax function. \\
In order to describe multiple obstacles at different known positions, we have to transform the ball's world coordinates \( \va{r}^{(\mathrm{W})} \) into the object coordinates \( \va{r}^{(\mathrm{o})} \) for each obstacle separately such that the obstacle is centered at the origin of the object coordinate system. This transformation can be realized using rotation and translation and is described by the transformation \( \va{r}^{(o)} = \mathcal{T}_{o} (\va{r}^{(\mathrm{W})} ) \) for each obstacle \( o \). As a result, the potential \( V_\mathrm{O} \) describing collisions with all obstacles in the specific environment is described as 
\begin{align}
    V_\mathrm{O}(\va{r}^{(\mathrm{W})}) =   \sum\limits_{o \in \mathcal{O}} V_{o}\left(\mathcal{T}_{o}(\va{r}^{(\mathrm{W})})\right)
\end{align}
with $\mathcal{O}$ being the set of all obstacles placed in the environment. \\[0.75ex]
With the potentials \( V_\mathrm{G} \), \( V_\mathrm{F} \), \( V_\mathrm{W} \) and \( V_\mathrm{O} \) defined, the differential equations of motion \ref{eq:hamiltonequations} can now be derived. This can be achieved using automatic differentiation, but we decided to calculate the derivatives of the Hamilton function analytically for the sake of a small speed up. We approximate the derivative of the ReLU function with a sigmoid function as 
\(
  \frac{\partial}{\partial x} \text{ReLU}(x) \approx \sigma (\beta_\text{g/w/f} \cdot x) 
\) 
with \( \beta_\text{g/w/f} = 200 \) being a large constant. There are no walls in the \textit{synthetic dataset} and, thus, we obtain the differential equations of motion as
\begin{equation}
    \resizebox{0.93\linewidth}{!}{$
  \begin{split}
    \frac{\mathrm{d}}{\mathrm{dt}} \va{r}^{(\mathrm{W})} &= \va{v} \text{ ,  } \\
    \frac{\mathrm{d}}{\mathrm{dt}} \va{v} &= - \begin{pmatrix} 0 \\ 0 \\ g \end{pmatrix} \sigma \left( \beta_\mathrm{g/w/f} \va{r}^{(\mathrm{W})}\right)
    + \begin{pmatrix} 0 \\ 0 \\ c \end{pmatrix} \sigma \left( - \beta_\mathrm{g/w/f} \va{r}^{(\mathrm{W})}\right) \\
    &+ \sum\limits_{o \in \mathcal{O}} \beta_o V_o(\va{r}^{(o)})  \mathcal{T}_o \\
    &\cdot \left( 1 - \sigma \left( \beta_o \left( \va{r}^{(o)} + \begin{pmatrix} l \\ w \\ 0 \end{pmatrix} \right)  \right) - \sigma \left( \beta_o  \left( \va{r}^{(o)} - \begin{pmatrix} l \\ w \\ h \end{pmatrix} \right)  \right)  \right) \text{.}
  \end{split}
  $}
\end{equation}
For the \textit{real dataset} we additionally consider the two walls and obtain the equations of motion as
\begin{equation}
  \resizebox{\linewidth}{!}{$
  \begin{split}
    \frac{\mathrm{d}}{\mathrm{dt}} \va{r}^{(\mathrm{W})} &= \va{v} \text{ ,  } \\
    \frac{\mathrm{d}}{\mathrm{dt}} \va{v} &= - \begin{pmatrix} 0 \\ 0 \\ g \end{pmatrix} \sigma \left( \beta_\mathrm{g/w/f} \va{r}^{(\mathrm{W})}\right)
    + \begin{pmatrix} 0 \\ 0 \\ c \end{pmatrix} \sigma \left( - \beta_\mathrm{g/w/f} \va{r}^{(\mathrm{W})}\right) \\
    &+ \begin{pmatrix} c \\ 0 \\ 0 \end{pmatrix} \sigma \left( - \beta_\mathrm{g/w/f} \va{r}^{(\mathrm{W})}\right) 
    + \begin{pmatrix} 0 \\ c \\ 0 \end{pmatrix} \sigma \left( \beta_\mathrm{g/w/f} \left( \va{r}^{(\mathrm{W})} - \SI{0.6}{m} \right) \right) \\
    &+  \sum\limits_{o \in \mathcal{O}} \beta_o V_o(\va{r}^{(o)})  \mathcal{T}_o \\
    &\cdot \left( 1 - \sigma \left( \beta_o \left( \va{r}^{(o)} + \begin{pmatrix} l \\ w \\ 0 \end{pmatrix} \right)  \right) - \sigma \left( \beta_o \left( \va{r}^{(o)} - \begin{pmatrix} l \\ w \\ h \end{pmatrix} \right)  \right)  \right) \text{.}
  \end{split}
  $}
\end{equation} \\[0.75ex]
In this paper we model the equations of motion directly. However, it is also possible to use a differential physics engine like BRAX \cite{brax} in the PAF. This would also allow to describe some physical phenomenons like spin even more easily.

\subsection{Obtaining the Camera Matrices}
We need the intrinsic as well as extrinsic camera matrix for our method. The intrinsic camera matrix is defined as 
\begin{align}
  \mathbf{M}_{int} =
  \begin{pmatrix}
    f_x & 0 & c_x \\
    0 & f_y & c_y \\
    0 & 0 & 1
  \end{pmatrix}
\end{align}
with \( f_{x\text{/}y} \) being the focal length and \( c_{x\text{/}y} \) being the principal point. With this matrix the \textit{euclidean camera coordinates} can be converted into \textit{homogeneous image coordinates}. The extrinsic camera matrix is defined as
\begin{align}
  \mathbf{M}_{ext} =
  \begin{pmatrix}
    \mathbf{R} & \mathbf{t} \\
    \mathbf{0}_{1 \times 3} & 1
  \end{pmatrix}
\end{align}
with \( \mathbf{R} \in \mathbb{R}^{3 \times 3} \) describing the rotation and \( \mathbf{t} \in \mathbb{R}^{3 \times 1} \) describing the translation of the world coordinate system. With this matrix the \textit{homogeneous world coordinates} are converted into \textit{homogeneous camera coordinates}. \\[0.75ex]
While the camera matrices are commonly available in modern broadcasting cameras \cite{DeepSportRadarv1}, they can also be calculated, for example, by using a Direct Linear Transform \cite{DLT}. This is done particularly straightforward in the context of sports, because characteristic points on the court can be easily annotated automatically. \\[0.75ex]
In the \textit{synthetic dataset} the camera matrices are provided by the physics engine MuJoCo, thus, the exact matrices are already available. In the \textit{real dataset} the camera is calibrated such that intrinsic camera matrix is already known. Therefore, we only have to calculate the extrinsic camera matrix. We label 11 characteristic points of the scene for each camera location and calculate an initial guess of the extrinsic camera matrix using the Direct Linear Transform. We then further optimize the extrinsic camera matrix using the BFGS optimization algorithm \cite{BFGS}. This way we obtain an accurate estimation of the extrinsic camera matrix. As this estimation is accurate but not exact, we conclude that our model is able to cope with approximate camera locations which is an important property for practical applications.

\begin{figure*}[th]
  \centering
  \includegraphics[width=0.85\linewidth]{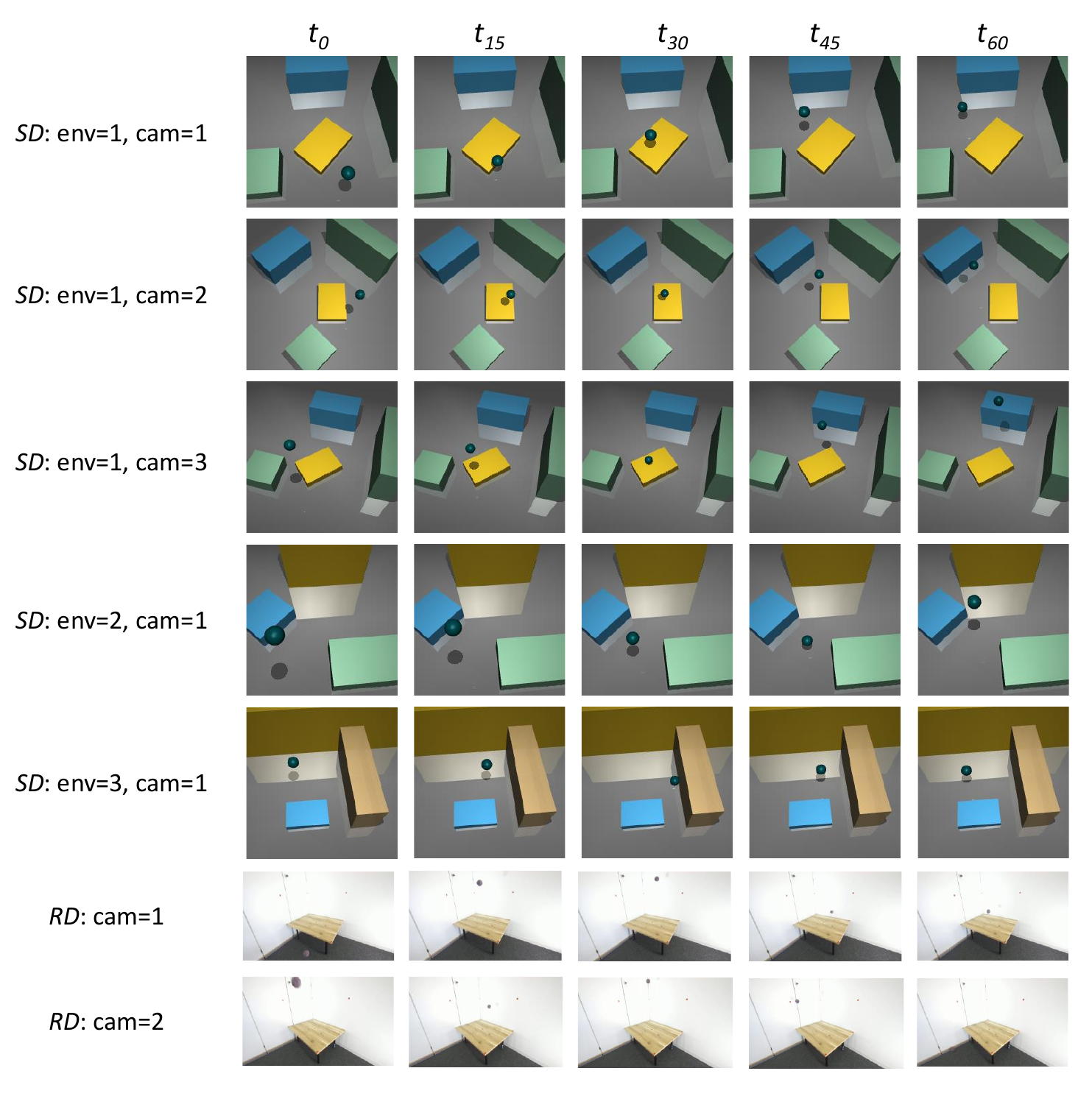}
  \caption{Frames from example sequences of multiple camera locations and environments are shown.}
  \label{img:datasets}
\end{figure*}

\section{Dataset}
\begin{figure*}[th]
    \centering
    \begin{minipage}[l]{0.45\linewidth}
      \captionsetup{width=0.9\linewidth}
      \captionof{table}{Camera locations in the synthetic dataset. Each camera looks at the origin of the world-coordinate system and is positioned at a radial distance $d$, polar angle $\theta$ and azimuthal angle $\phi$ with respect to the origin.}
      \centering
      \begin{tabular}{ c c c c }
        \toprule
        camera & $d$ (m) & $\theta$ (°) & $\phi$ (°) \\
        \midrule
        1 & $8$ & $-60$ & $40$ \\ 
        2 & $10$ & $-65$ & $0$ \\  
        3 & $10$ & $-55$ & $60$ \\ 
        4 & $9$ & $-60$ & $20$ \\  
        5 & $7$ & $-70$ & $50$ \\ 
        6 & $9$ & $-50$ & $10$ \\  
        7 & $10$ & $-60$ & $30$ \\ 
        8 & $8$ & $-70$ & $0$ \\  
        9 & $8$ & $-50$ & $60$ \\ 
        \bottomrule
      \end{tabular}
      \label{tab:cameraparamter}
    \end{minipage}%
    \hfill
    \begin{minipage}[r]{0.45\linewidth}
      \centering
      \includegraphics[width=0.95\linewidth]{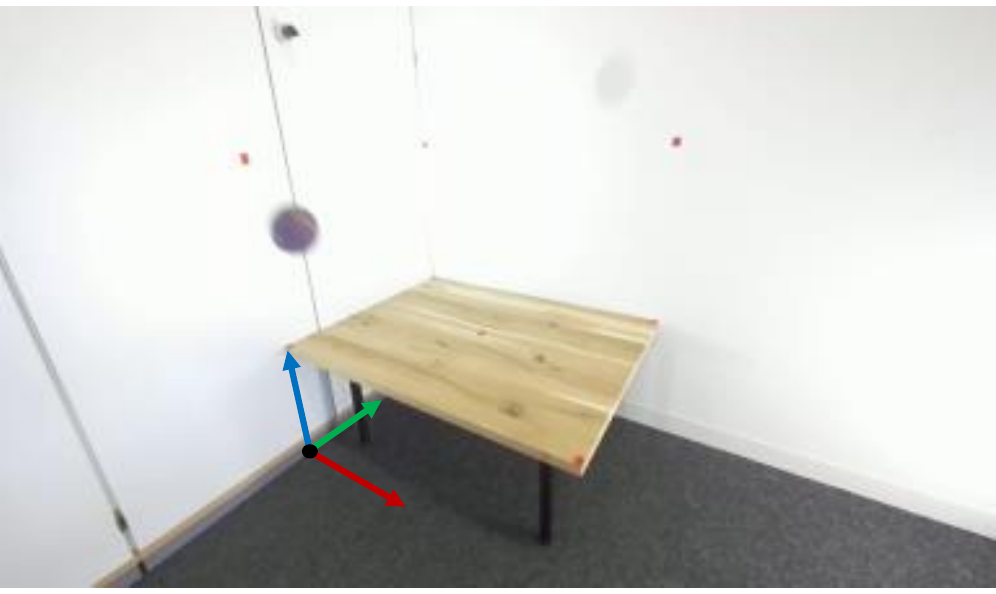}
      \captionof{figure}{Origin of the \textit{real dataset}. Blue represents the z-axis, green the y-axis and red the x-axis.}
      \label{img:origin}
    \end{minipage}
  \end{figure*}

  In this section, we present additional details on the datasets introduced in the main paper. In Figure \ref{img:datasets}, we visualize some exemplary videos from different camera positions and in different environments to provide a better overview of the datasets. In the \textit{synthetic dataset}, we record the videos from 9 possible camera locations that are shown in Table \ref{tab:cameraparamter}. All camera locations are chosen such that the origin of the world coordinate system corresponds to the center of the image. We set the gravitational constant to \( g = \SI{1}{\frac{m}{s^2}} \). This reduces the speed of the ball, making it easier to automatically generate the synthetic dataset, because we have to ensure that the ball does not leave the field of view. \\[0.75ex]
  Since we record all videos in the \textit{real dataset} with a ZED 2i Stereo Camera, we are able to calculate a ground truth depthmap for each image. For this purpose, we select the \textit{Neural Depth Mode} of the camera to obtain the best possible results. Nevertheless, the results are still noisy and, thus, we extract the ground truth depth at the annotated ball's position as average over the neighboring values of the depth map \( D \) as
  \begin{align}
    z^{(\mathrm{C})} = \frac{1}{9} \cdot \sum\limits_{i = -1}^{1} \sum\limits_{j = -1}^{1} D_{x^{(\mathrm{I})} + i,\, y^{(\mathrm{I})} + j} 
  \end{align}
  with \( \begin{pmatrix} x^{(\mathrm{I})} & y^{(\mathrm{I})} \end{pmatrix}^\mathrm{T} \) being the ground truth ball's position in image coordinates. Based on this ground truth depth and the intrinsic camera matrix we calculate the ground truth camera coordinates. We define a fixed origin of the world coordinate system such that the physical potentials can be described easily, and we visualize the origin in Figure \ref{img:origin}. The 2D image coordinates are annotated manually by simply clicking on the ball's position in the image. To expedite this process, we use a simple automatic ball detection algorithm based on the Hough Transformation and only correct the ball position manually if the automatic detection fails.

\section{Experiments}
In this section we give further details on the training process, provide additional results from the main paper and discuss new experiments.

\subsection*{Discussion of training process} 
Our code is implemented in PyTorch \cite{pytorch}, and we utilize models and architectures from the timm library \cite{timm}. We optimize our model with the ADAM \cite{adam} optimizer and use a batch size of \( 8 \) together with a learning rate of \( 2 \cdot 10^{-5} \). The Dormand-Prince method \cite{dopri5} is applied for the numerical solution of the differential equations \ref{eq:hamiltonequations} and its implementation utilizes the torchdyn \cite{torchdyn} library. \\
We train the \textit{RD} and \textit{SD-S} models for 400 epochs, the \textit{SD-M} model for 200 epochs and the \textit{SD-L} model for 100 epochs. Although the models show slight improvements with longer training, the performance gains are small. We evaluate the model after each epoch on the validation set and select the model with the best performance for the final evaluation on the test set. We multiply the \textit{future loss} (first term in equation \ref{eq:loss}) with a factor, which linearly increases from 0 to 1 during the first 300 epochs of the training. This way we ensure that the model focuses on learning to recognize the ball in the images first, and thus, prevents a model collapse. \\
Our experiments were conducted on a combination of Nvidia RTX 3090, V100, A100, and H100 GPUs. Therefore, we do not provide the actual runtime of our experiments as a comparison is not meaningful. However, to provide a general idea of the runtime, training our model for 400 epochs on the \textit{real dataset} in Table \ref{tab:DtG_real} takes approximately 12 hours on a single V100 GPU. We note that a considerable good performance is already reached after only 100 epochs (3 hours on a single V100 GPU). Furthermore, we anticipate that further code optimizations, such as improved data loading techniques or model compilation, could lead to significant increases in training speed. For all results in this paper we report the mean and standard deviation (represented by the symbol \( \Delta \)) of the metric calculated across all images in the test set.\\
While we provide only results for camera locations 1, 7, 8, and 9 in Table \ref{tab:DtG_synthetic_cam}, the results for all camera locations are given in Table \ref{tab:DtG_synthetic_cam_full}. We see that the \textit{SD-M} and \textit{SD-L} models are able to estimate the ball's 3D position for all camera locations very precisely. Since the \textit{SD-L} model achieves better scores for most camera locations, this reinforces our assumption that training with additional data recorded in multiple environments is actually beneficial.  

\begin{table*}[ht]
\small
\captionsetup{width=\linewidth}
\caption{\( \textit{DtG} \) scores per camera location evaluated on the 1\textsuperscript{st} environment of the \textit{synthetic dataset}. An extension of Table \ref{tab:DtG_synthetic_cam} from the main paper.}
\label{tab:DtG_synthetic_cam_full}
\centering
\begin{tabular}{cccccccccc}
  \toprule
    & \multicolumn{9}{c}{ \( \textit{DtG} \pm \Delta \textit{DtG} \) (cm) \(\downarrow\)} \\
  \cmidrule(r){2-10} 
  training set & camera 1 & camera 2 & camera 3 & camera 4 & camera 5 & camera 6 & camera 7 & camera 8 & camera 9 \\
  \midrule
  \textit{SD-S} & \( 22 \pm 19 \) & - & - & - & - & - & - & - & - \\
  \textit{SD-M}  & \( 19 \pm 10 \) & \( 19 \pm 18 \) & \( 28 \pm 24 \) & \( 13 \pm 9 \) & \( 26 \pm 12 \) & \( 19 \pm 14 \) & \( 27 \pm 23 \) & \( 23 \pm 9 \) & \( 21 \pm 10 \) \\
  \textit{SD-L} & \( 11 \pm 6 \) & \( 20 \pm 19 \) & \( 28 \pm 25 \) & \( 11 \pm 8 \) & \( 10 \pm 6 \) & \( 19 \pm 13 \) & \( 28 \pm 25 \) & \( 15 \pm 8 \) & \( 16 \pm 7 \) \\
  \bottomrule
\end{tabular}
\label{tab:DtG_synthetic_cam_full}
\normalsize
\end{table*}

\subsection*{Comparison of different backbones}
We perform an additional experiment to compare the performance of different backbones in the PEN. For each backbone, we use the implementation provided in the timm library. In the other experiments, we always use the first 4 out of 5 stages of a ResNet34. We now compare this backbone with 4 out of 4 stages of the ConvNeXt-nano architecture and 4 out of 4 stages of the ConvNeXtv2-nano architecture. All models are pretrained on ImageNet-1K, and we train and test them on the \textit{SD-M} dataset. The results are shown in table \ref{tab:DtG_backbones}. \\[0.75ex] \noindent 
\begin{table*}[ht]
  \small
  \captionsetup{width=\textwidth}
  \caption{Results for different backbones. Each model is trained on the \textit{SD-M} dataset. \( \textit{DtG} \) scores per camera location evaluated on the 1\textsuperscript{st} environment of the \textit{synthetic dataset}.}
  \label{tab:DtG_backbones}
  \centering
  \resizebox{\textwidth}{!}{
  \begin{tabular}{ccccccccccc}
    \toprule
      & & \multicolumn{9}{c}{ \( \textit{DtG} \pm \Delta \textit{DtG} \) (cm) \(\downarrow\)} \\
    \cmidrule(r){2-11} 
    backbone & \# params & camera 1 & camera 2 & camera 3 & camera 4 & camera 5 & camera 6 & camera 7 & camera 8 & camera 9 \\
    \midrule
    resnet & \( 0.8 \cdot 10^7 \) & \( 19 \pm 10 \) & \( 19 \pm 18 \) & \( 28 \pm 24 \) & \( 13 \pm 9 \) & \( 26 \pm 12 \) & \( 19 \pm 14 \) & \( 27 \pm 23 \) & \( 23 \pm 9 \) & \( 21 \pm 10 \)  \\
    convnext  & \( 1.5 \cdot 10^7 \) & \( 14 \pm 12 \) & \( 26 \pm 21 \) & \( 32 \pm 27 \) & \( 15 \pm 13 \) & \( 15 \pm 11 \) & \( 24 \pm 21 \) & \( 37 \pm 31 \) & \( 22 \pm 11 \) & \( 20 \pm 12 \) \\
    convnextv2 & \( 1.5 \cdot 10^7 \) & \( 23 \pm 14 \)  & \( 33 \pm 28 \) & \( 45 \pm 32 \) & \( 25 \pm 20 \) & \( 18 \pm 11 \) & \( 31 \pm 27 \) & \( 49 \pm 36 \) & \( 53 \pm 22 \) & \( 35 \pm 20 \) \\
    \bottomrule
  \end{tabular}
  }
  \normalsize
\end{table*}
Interestingly, the ConvNeXtv2 backbone performs notably worse than the other two architectures, despite its seemingly advanced design. In summary, the ResNet performs slightly better than the ConvNeXt, despite the ConvNeXt having twice as many parameters. One reason for this behavior might be that our input image resolution is small, and we do not benefit from a very large receptive field. Since the ConvNeXt architecture uses \( 7 \times 7 \) convolutions, the receptive field grows faster than in the ResNet. Once the receptive field size becomes as large as the image resolution, the additional stages may not provide significant benefits and could potentially hinder the model's ability to learn an effective representation. Since the ResNet only uses \( 3 \times 3 \) convolutions and the features are extracted at an earlier stage compared to the other architectures, the receptive field is smaller and the model might be able to learn a better representation. However, additional experiments using different backbones and higher input image resolutions are required to validate this hypothesis. We note that our code can be readily extended with other backbones, provided that they support the extraction of 4 feature stages, which is a requirement in our implementation. If a different number of features is used, the implementation of the \textit{depthmap head} and \textit{heatmap head} needs to be adjusted. In conclusion, even though the ResNet is a relatively old architecture, it remains a highly suitable choice for our task.

\subsection*{Analysis of future loss}
In this section, we delve deeper into the concept of the \textit{future loss}. As can be seen in figure \ref{img:overview}, we compare the image coordinates \( \va{r}^{(\mathrm{I})}_\text{PEN}(t_{n + \Delta n}) \) predicted by the PEN with the coordinates \( \va{r}^{(\mathrm{I})}_\text{PAF}(t_{n + \Delta n}) \) forecasted by the PAF. This way of computing the \textit{future loss} is explicitly written in the first term of equation \ref{eq:loss}, and we refer to this method as \textit{2D-predict} in this section. Because the ground truth image coordinates are available for every frame, we can also compute the \textit{future loss} by comparing the PEN's predictions with the ground truth coordinates \( \va{r}^{(\mathrm{I})}_\text{GT}(t_{n + \Delta n}) \). Consequently, we simply replace the coordinates \( \va{r}^{(\mathrm{I})}_\text{PAF}(t_{n + \Delta n}) \) with \( \va{r}^{(\mathrm{I})}_\text{GT}(t_{n + \Delta n}) \) in equation \ref{eq:loss}. This way of computing the loss is denoted as \textit{2D-gt}. Moreover, we also add a third method of computing the \textit{future loss} by comparing the PEN's predictions with the ground truth camera coordinates \( \va{r}^{(\mathrm{C})}_\text{GT}(t_{n + \Delta n}) \), which are usually not used for the training. Consequently, the \textit{future loss} is computed in this case as
\begin{align}
  \textit{future loss} = \norm{\va{r}^{(\mathrm{C})}_\text{PEN}(t_{n + \Delta n}) - \va{r}^{(\mathrm{C})}_\text{GT}(t_{n + \Delta n})}_\text{L1} \, .
\end{align}
We refer to this way of computing the loss as \textit{3D-gt} in this section, and we use this method as baseline. We note that for this way of computing the loss, we utilize the 3D ground truth, which is usually not available in real world scenarios. Nonetheless, we believe that examining this baseline can provide valuable insights into our analysis. \FloatBarrier\vspace{-\parskip + 0.75ex}
\begin{table}[ht]
  \small
  \captionsetup{width=\linewidth}
  \caption{\( \textit{DtG} \) scores and \( \textit{DtG}_3 \) scores per camera location for the \textit{real dataset}. Models trained with different \textit{future losses} are compared.}
  \label{tab:DtG_loss}
  \centering
  \begin{tabular}{ccccc}
    \toprule
      & \multicolumn{2}{c}{ \( \textit{DtG} \pm \Delta \textit{DtG} \) (cm) \(\downarrow\)} & \multicolumn{2}{c}{ \( \textit{DtG}_3 \pm \Delta \textit{DtG}_3 \) (cm) \(\downarrow\)} \\
    \cmidrule(r){2-5} 
    \textit{future loss} & camera 1 & camera 2 & camera 1 & camera 2 \\
    \midrule
    \textit{2D-predict} & \( 7 \pm 4 \) & \( 6 \pm 4 \) & \( \begin{pmatrix} 8 \pm 4 \\ 6 \pm 2 \\ 11 \pm 4 \end{pmatrix} \) & \( \begin{pmatrix} 4 \pm 2 \\ 6 \pm 3 \\ 11 \pm 4 \end{pmatrix} \) \\
    \textit{2D-gt} & \( 18 \pm 10 \) & \( 17 \pm 10 \) & \( \begin{pmatrix} 6 \pm 5 \\ 14 \pm 7 \\ 33 \pm 7 \end{pmatrix} \) & \( \begin{pmatrix} 5 \pm 3 \\ 15 \pm 6 \\ 32 \pm 7 \end{pmatrix} \) \\
    \textit{3D-gt} & \( 3 \pm 3 \)  & \( 3 \pm 4 \) & \( \begin{pmatrix} 3 \pm 2 \\ 2 \pm 2 \\ 6 \pm 5 \end{pmatrix} \) & \( \begin{pmatrix} 3 \pm 2 \\ 2 \pm 3 \\ 6 \pm 5 \end{pmatrix} \) \\
    \bottomrule
  \end{tabular}
  \normalsize
\end{table}
We train three models with the different future losses and report the results in table \ref{tab:DtG_loss}. By comparing the \textit{2D-predict} model with the \textit{2D-gt} model, we can see that the \textit{2D-predict} model performs better than the \textit{2D-gt} model. This is due to the ground truth image coordinates being noisy, since they are annotated manually. Obviously, the PEN learns to predict more accurate image coordinates and, consequently, the calculation of the \textit{future loss} is more accurate. \\
As expected, the \textit{3D-gt} model performs best, since it is trained with 3D ground truth information. However, despite the \textit{2D-predict} model being worse, it is still able to predict the 3D position of the ball surprisingly well, since the metric is in the same order of magnitude as with the \textit{3D-gt} model. Consequently, we conclude that the method presented in this paper provides a viable alternative to fully supervised training. Additionally, we highlight the potential for combining our method with supervised training, particularly in scenarios where 3D ground truth data is only accessible for a subset of the data. Such integration could lead to further enhancements in existing applications. 

\subsection*{Learning arbitrary physics}
\begin{figure*}[t]
  \centering
  \includegraphics[width=0.99\textwidth]{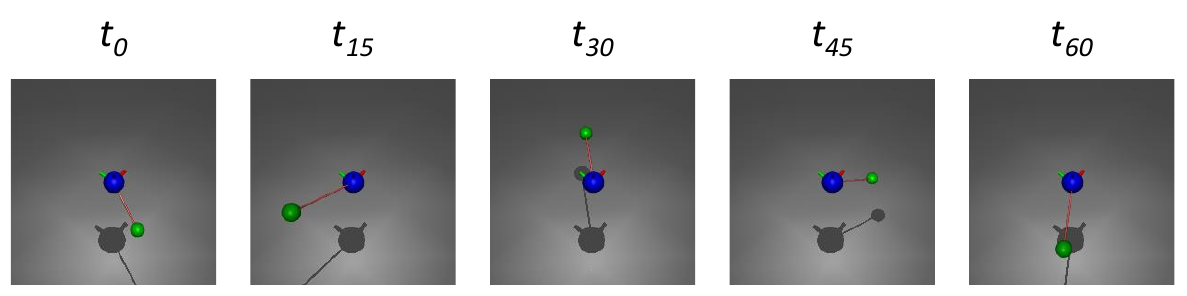}
  \caption{Frames from an example video of the \textit{spring dataset}.}
  \label{img:dataset_carousel}
\end{figure*}

An important advantage of our method is its ability to describe arbitrary physical systems by utilizing numerical solutions for the equations of motion, rather than relying on analytical solutions. In this section we show two things:
\begin{itemize}[leftmargin=0.5cm,itemsep=0pt, topsep=-0.5\parskip]
  \item We demonstrate that our method can also be used for physical systems that behave differently from the previous described bouncing balls.
  \item We calculate an analytic solution of the equations of motion and compare the numerical forecasting in the PAF with the analytic forecasting.
\end{itemize}
Therefore, we introduce a new synthetic dataset denoted as \textit{spring dataset}. It consists of video of a moving ball connected with a spring to a fixed ball in the center. The moving ball of mass \( m=\SI{1}{kg} \) is attracted to the fixed center ball by the spring with spring constant \( k = \SI{3}{\frac{N}{m}} \). We record 100 videos from a single fixed camera location for training, 100 videos for validation and 100 videos for testing. The resolution of the frames is \( 224 \times 224 \) and the videos are recorded with \( \SI{30}{\fps} \). We visualize frames from an example video depicting the motion of the ball connected to the spring in Figure \ref{img:dataset_carousel}. \\[0.75ex]
According to Hooke's law, the potential of the moving ball is given by \( V(\va{r}^{(\mathrm{W})}) = \frac{1}{2} k  \left| \va{r}^{(\mathrm{W})} \right| ^2 \) and by applying the Hamilton formalism (see equation \ref{eq:hamiltonequations}), we obtain the equations of motion as
\begin{align}
  \label{eq:carousel_numeric}
  \frac{\mathrm{d}}{\mathrm{dt}} \va{r}^{(\mathrm{W})} = \va{v} \hspace{0.2cm} \text{, } \hspace{0.3cm} \frac{\mathrm{d}}{\mathrm{dt}} \va{v} = - \frac{k}{m} \cdot \va{r}^{(\mathrm{W})} \hspace{0.15cm} \,.
\end{align}
The analytic solution to these differential equations can be experessed as
\begin{align}
  \label{eq:carousel_analytic}
  \va{r}^{(\mathrm{W})}(t) = \frac{1}{k} \cdot \va{v}_0 \cdot \sin \left( k \cdot t \right) + \va{r}^{(\mathrm{W})}_0
\end{align}
where \( \va{r}^{(\mathrm{W})}_0 \) and \( \va{v}_0 \) are the initial position and velocity of the moving ball. \\[0.75ex]
We implement the numeric solver of equation \ref{eq:carousel_numeric} similar to the previous experiments, and compare its performance with the utilization of the analytic function \ref{eq:carousel_analytic} in the PAF. The results are given in Table \ref{tab:DtG_carousel}. 
\begin{table}[ht]
  \small
  \captionsetup{width=\linewidth}
  \caption{\( \textit{DtG} \) and \( \textit{DtG}_3 \) scores on the \textit{spring dataset}. The analytic solution of the equations of motion is compared to a numeric solution.}
  \label{tab:DtG_carousel}
  \centering
  \begin{tabular}{ccc}
    \toprule
    PAF mode & \( \textit{DtG} \pm \Delta \textit{DtG} \) (cm) \(\downarrow\) & \( \textit{DtG}_3 \pm \Delta \textit{DtG}_3 \) (cm) \(\downarrow\) \\
    \midrule
    numeric & \( 9 \pm 15 \) & \( \begin{pmatrix} 6 \pm 6 \\ 5 \pm 6 \\ 16 \pm 22 \end{pmatrix} \)  \\
    analytic & \( 10 \pm 16 \) & \( \begin{pmatrix} 8 \pm 8 \\ 4 \pm 5 \\ 21 \pm 21 \end{pmatrix} \)  \\
    \bottomrule
  \end{tabular}
  \normalsize
\end{table}
The model trained with the numeric PAF and the model trained with the analytical PAF perform very similarly, with the numeric version achieving slightly better scores. This shows that using numeric solvers in the PAF is a valid approach that is able to describe arbitrary physical systems. \\
In general, the numeric model achieves a \( \textit{DtG} \) of just \( \SI{9}{cm} \), demonstrating the applicability of our method to physical systems beyond bouncing balls. \\[0.75ex] 
While the physics of all datasets presented in this paper is determined by the laws of classical mechanics, it is worth considering the potential extension of our method to other domains such as quantum mechanics or electrodynamics, where analytic solutions are not available in most cases. However, exploring these possibilities lies beyond the scope of this work.

\subsection*{Discussion of stability}
To ensure that our results are not just random fluctuations, we test our model with different random seeds and compare the results. For this experiment we choose the \textit{SD-S} scenario from the main paper and set 8 different random seeds at the beginning of the training. This results in a different initialization of the \textit{heatmap head} and \textit{depthmap head}, as well as in a different ordering of the training data in each epoch. We present the results of the models in Table \ref{tab:DtG_seeds}. \FloatBarrier\vspace{0.75ex}
\begin{table}
  \small
  \captionsetup{width=0.5\textwidth}
  \caption{\( \textit{DtG} \) scores for the 1\textsuperscript{st} camera location in the 1\textsuperscript{st} environment. The results of the \textit{SD-S} model are compared for different initial seeds.}
  \label{tab:DtG_seeds}
  \centering
  \begin{tabular}{cc}
    \toprule
    seed & \( \textit{DtG} \pm \Delta \textit{DtG} \) (cm) \(\downarrow\) \\
    \midrule
    1 & \( 24 \pm 20 \)  \\
    2 & \( 27 \pm 21 \)  \\
    3 & \( 26 \pm 21 \)  \\
    4 & \( 23 \pm 19 \)  \\
    5 & \( 23 \pm 21 \)  \\
    6 & \( 25 \pm 21 \)  \\
    7 & \( 25 \pm 19 \)  \\
    8 & \( 22 \pm 21 \)  \\
    \bottomrule
  \end{tabular}
  \normalsize
\end{table}
The results obtained for all seed values show a high degree of similarity. The mean value across multiple runs is \( E_\textit{DtG} = \SI{24.4}{cm} \), with a small standard deviation of \( \sigma_\textit{DtG} = \SI{1.6}{cm} \). The low standard deviation among individual runs suggests that the outcomes are not mere random fluctuations. This observation is further supported by the fact that the standard deviation over the multiple runs is an order of magnitude smaller than the standard deviations among the individual images \( \Delta \textit{DtG} \). Hence, the effect of the random seed is negligible. \\[0.75ex]
While the results are relatively stable over different seeds, we observe that other factors have a much larger impact on the results. One such factor is the selection of the number of forecast steps. As described in equation \ref{eq:loss}, we compare the coordinates at time \( t_{n + \Delta n} \) with the forecasted coordinates at this time. Depending on the speed of the ball, a good value for \( \Delta n \) has to be chosen carefully. If \( \Delta n \) is too small, the ball may not have moved a significant distance, resulting in a very low loss. Consequently, the PEN may converge to a trivial solution. Conversely, if \( \Delta n \) is too large, the distance might be too large, and the PEN may struggle to learn the correct solution. Therefore, for each scenario an appropriate value for \( \Delta n \) has to be chosen. \\
In this paper, we do not use a single value for \( \Delta n \), instead we calculate the \textit{future loss} for multiple values of \( \Delta n \) and then compute the average. This way, the small \( \Delta n \) values help to stabilize the training, while larger ones ensure that the PEN does not collapse to a trivial solution. Although this approach aids in training, it is still necessary to choose appropriate values for \( \Delta n \). \\[0.75ex]
One way to significantly improve the stability of the training is to use a better pretraining of the PEN. For example a segmentation or even a depth estimation task could be chosen for pretraining the model. Since this teaches the model 3D knowledge, a model collapse is less likely to happen. However, in this paper we want to show that our method is able to learn 3D dynamics without any 3D supervision. Thus, we do not further explore more advanced pretraining strategies, and instead only initialize  the backbone of the PEN with ImageNet weights in our experiments.

\FloatBarrier
\subsection*{Limitations and improvements}
We view this paper as a step towards 3D object location estimation without requiring 3D supervision. Our findings demonstrate the feasibility of training a model without depth information and provide a comprehensive analysis of our approach. However, there are still several limitations to address. Specifically, we only use low resolution images to ensure an efficient training of our models. It is possible to use such low resolution images in our experiments, because the ball's relative size in the images is large enough to be clearly identified. In contrast, sports videos typically involve high-resolution frames where the ball appears much smaller. Consequently, simply downscaling the entire image would not suffice to maintain clear ball identification. Hence, the development of more sophisticated methods is necessary to optimize computation time. One approach is to leverage an object detection pipeline to extract the relevant region around the ball, which can then be utilized for 3D location estimation. We anticipate that the insights presented in this paper will inspire the advancement of more sophisticated techniques in this field.